# Generative networks as inverse problems with fractional wavelet scattering networks

Jiasong Wu[1, 2, 3], Jing Zhang[1, 3], Fuzhi Wu[1, 3], Youyong Kong[1, 3],

Guanyu Yang[1, 3], Lotfi Senhadji[2, 3], Huazhong Shu[1, 3]

[1]*Laboratory of Image Science and Technology, Key Laboratory of Computer Network and Information Integration, Southeast University, Ministry of Education, Nanjing 210096, China*

[2]*Université de Rennes, INSERM, LTSI-UMR 1099, Rennes F-35042, France*

[3]*Centre de Recherche en Information Biomédicale Sino-français (CRIBs), Université de Rennes, INSERM, Southeast University, Rennes, France, Nanjing, China*

**Abstract**：Deep learning is a hot research topic in the field of machine learning methods and applications. Generative Adversarial Networks (GANs) and Variational Auto-Encoders (VAEs) provide impressive image generations from Gaussian white noise, but both of them are difficult to train since they need to train the generator (or encoder) and the discriminator (or decoder) simultaneously, which is easy to cause unstable training. In order to solve or alleviate the synchronous training difficult problems of GANs and VAEs, recently, researchers propose Generative Scattering Networks (GSNs), which use wavelet scattering networks (ScatNets) as the encoder to obtain the features (or ScatNet embeddings) and convolutional neural networks (CNNs) as the decoder to generate the image. The advantage of GSNs is the parameters of ScatNets are not needed to learn, and the disadvantage of GSNs is that the expression ability of ScatNets is slightly weaker than CNNs and the dimensional reduction method of Principal Component Analysis (PCA) is easy to lead overfitting in the training of GSNs, and therefore affect the generated quality in the testing process. In order to further improve the quality of generated images while keep the advantages of GSNs, this paper proposes Generative Fractional Scattering Networks (GFRSNs), which use more expressive fractional wavelet scattering networks (FrScatNets) instead of ScatNets as the encoder to obtain the features (or FrScatNet embeddings) and use the similar CNNs of GSNs as the decoder to generate the image. Additionally, this paper develops a new dimensional reduction method named Feature-Map Fusion (FMF) instead of PCA for better keeping the information of FrScatNets and the effect of image fusion on the quality of image generation is also discussed. The experimental results on CIFAR-10 dataset and CelebA dataset show that the proposed GFRSNs can obtain better generated images than the original GSNs on the testing dataset.

# 1. Introduction

Generative models attract the attention of many researchers recently and are widely used in image synthesis, image restoration, image inpainting, image reconstruction, etc. Many generative models have been constructed in the literature. These models can be roughly classified into two kinds [1]: explicit density methods and implicit density methods.

Among explicit density generative models, Variational Auto-Encoders (VAEs) [2] and their variants [3-20] are probably the most commonly used models since they have useful latent representation, which can be used in inference queries. Kingma and Welling [2] first proposed VAEs, which train the encoder and decoder simultaneously and can perform efficient inference and learning in directed probabilistic models and in the presence of continuous latent variables with intractable posterior distributions. Salimans et al. [4] bridged the gap between Markov Chain Monte Carlo (MCMC) and VAE, and incorporated one or more steps of MCMC into variational approximation. Sohn et al. [5] proposed Conditional VAE (CVAE), which joins existing label information in training to generate corresponding category data. Rezende and Mohamed [6] introduced a new approach for specifying flexible, arbitrarily complex and scalable approximate posterior distributions and provided a clear improvement in performance and applicability of variational inference. Sønderby et al. [7] presented a Ladder Variational Autoencoder, which recursively corrects the generative distribution by a data dependent approximate likelihood in a process resembling the Ladder Network. Shang et al. [8] proposed channel-recurrent variational autoencoders (CR-VAE), which integrates recurrent connections across channels to both inference and generation steps of VAE and generates face and bird images with high visual quality. Higgins et al. [10] presented a $β$-VAE, which is a modification of Variational Autoencoder (VAE) with a special emphasis to discover disentangled latent factors. Den Oord et al. [11] proposed a simple yet powerful generative model that learns discrete representations and allowed the model to circumvent issues of posterior collapse. Gregor et al. [12] proposed Temporal Difference VAE (TD-VAE), which is a generative sequence model that learns representations containing explicit beliefs about states several steps into the future. Razavi et al. [13] proposed Vector Quantized Variational AutoEncoder (VQ-VAE), which augments with powerful priors over the latent codes and is able to generate samples with quality that rivals that of state of the art GANs on multifaceted datasets such as ImageNet. Simonovsky and Komodakis [17] proposed Graph VAE which sidesteps hurdles of linearization of discrete structures by outputting a probabilistic fully-connected graph of a predefined maximum size directly at once. Please refer to [18, 19] for more references on VAEs.

Among implicit density generative models, Generative Adversarial Networks (GANs) [20] and their variants [21-62] are probably the most commonly used models since they have better generated images than other generative models. Goodfellow et al. [20] first proposed GANs, which estimating generative models via an adversarial process where a generative model G and a discriminative model D are trained simultaneously and there is no need for Markov chains or unrolled approximate inference networks during either training or generation of samples. However, applying GANs to real-world computer vision problems still encounter at least three significant challenges [21]: (1) High quality image generation; (2) Diverse image generation; and (3) Stable training. Then, many variants of GAN have been constructed in the literature to handle these three challenges. These variants of GAN can be roughly classified into two kinds [21]: architecture variant GANs [22-36] and loss variant GANs [37-50].

In terms of architecture variant GANs, for example, Radford et al. [22] proposed Deep Convolutional GAN (DCGAN), which uses a convolutional neural network (CNN) as the discriminator D and deploys deconvolutional neural network architecture for G and the spatial upsampling ability of the deconvolution operation enables the generation of higher resolution images compared to the original GANs [20]. Mirza and Osindero [23] proposed Conditional GAN (CGAN) which imposes a condition of additional information such as a class label to control the data generation process in a supervised manner. CGAN is then extended to Auxiliary classifier GAN (AC-GAN) [24] and Plug and play generative networks (PPGN) [25]. Chen et al. [26] presented InfoGAN which decomposes an input noise vector into a standard incompressible latent vector and another latent variable to capture salient semantic features of real samples. Karras et al. [33] presented progressive GAN for generative high-resolution images using the idea of progressive neural networks [34], which does not suffer from forgetting and is able to deploy prior knowledge via lateral connections to previously learned features.

In terms of loss-variant GANs, for example, Arjovsky et al. [37] proposed Wasserstein GAN (WGAN), which use the Wasserstein distance as the loss measure for optimization instead of Kullback-Leibler divergence. Gulrajani et al. [38] proposed an improved method for training the discriminator for a WGAN, by penalizing the norm of discriminator gradients with respect to data samples during training, rather than performing parameter clipping. Nowozin et al. [43] proposed an alternative cost which is function f-divergence for updating the generator which is less likely to saturate at the beginning of training. Qi [45] presented Loss Sensitive GAN (LS-GAN), which train the generator to produce realistic samples by minimizing the designated margins between real and generated samples. Miyato et al. [49] proposed Spectral Normalization GAN (SN-GAN), which use weight normalization technique to train the discriminator more stably. Brock et al. [50] proposed BigGAN, which use hinge loss instead of Jensen-Shannon divergence and a large-scale dataset to train the generator to produce more realistic samples. Please refer to [21, 51, 52] for more references on GANs and refer to [53-62] for some recent applications of GANs.

Although GANs and VAEs are great generative models, they raise many questions. A significant disadvantage of VAEs is that the resulting generative models produce blurred images compared with GANs since the quality of the VAEs crucially relies on the expressiveness of their inference models. A significant disadvantage of GANs [63-66] is that the training process is very difficult and may lead to unstable training and also model collapse. *Thus, a first question raised*: *Can we design a network that can maintain the characteristics of high-quality generated images of GANs as much as possible while reduce the training difficulty of GANs?* To handle this problem, Angles and Mallat proposed Generative Scattering Networks (GSNs) [66], which use wavelet scattering networks (ScatNets) [67, 68] as the encoder to obtain the features (or ScatNet embeddings) and the deconvolutional neural network of DCGAN [22] as the decoder to generate the image. The advantage of GSNs is the parameters of ScatNets are not needed to learn and therefore reduce the training difficulty when compared to DC-GAN, and the disadvantage of GSNs is that the generated images are easily to lose details and affect the image generated quality. After carefully inspection, we find that the sources of relative low-quality generated images of GSNs at least include two aspects: (*a*) the expression ability of ScatNets is slightly weaker than the CNNs used in DC-GAN; (*b*) applying the Principal Component Analysis [69] (PCA) to reduce the dimension of the feature map of ScatNets in the encoder part of GSNs leads to overfitting problem in the testing process of GSNs. *Then, a second question raised: Can we change the method of feature extraction of ScatNets to other more powerful one that is also not need learning? Can we develop a more suitable dimensional reduction method to solve the overfitting problem in the testing process of GSNs?*

In an attempt to solving the second question, in this paper, we propose Generative Fractional Scattering Networks (GFRSNs) which can be seen as an extension of GSNs. The contributions of the paper are as follows:

1) We use more expressive fractional wavelet scattering networks (FrScatNets) [70] instead of ScatNets) [67, 68] to extract the features of images and use image fusion [71, 72] in GFRSNs to effectively improve the visual quality of generated images.

2) We proposed a new dimensional reduction method named Feature-Map Fusion (FMF) which is more suitable for reducing the feature dimension of FrScatNets than PCA since the FMF method greatly alleviates the overfitting problem on testing datasets by using GFRSNs.

3) The generated testing images from the proposed GFRSNs on CIFAR-10 and CelebA datasets are better than the original GSNs.

The rest of the paper is organized as follows. In Section 2, wavelet scattering networks and the architectural components of GSNs are briefly introduced. The main architectural components of GFRSNs which include fractional wavelet scattering networks, FMF dimensional reduction method and generative networks and an image fusion method are introduced in Section 3. The performance of generated image of GFRSNs is analyzed and also compared to the original GSNs in Section 4. The conclusions are formulated in Section 5. Table 1 lists the fundamental symbols defined in this paper.

Table 1 List of symbols

| Symbol | Description |
|---|---|
| $\mathcal{X}$ | The input images |
| $\tilde{\mathcal{X}}$ | The generated images with PCA based GSNs |
| $\mathcal{X}^{(i)}$ | The $i$-th input image, $i=1,2,...,M$ |
| $\tilde{\mathcal{X}}^{(i)}$ | The $i$-th generated sample with PCA based GSNs, $i=1,2,...,M$ |
| $\tilde{\mathcal{X}}_\alpha$ | The generated images with FMF based GFRSNs with fractional order $\alpha$ |
| $\mathcal{Y}$ | The ouput features from ScatNet (ScatNet embeddings) |
| $\mathcal{Y}_\alpha$ | The ouput features from FrScatNet with fractional order $\alpha$ (FrScatNet embeddings) |
| $\mathbf{z}$ | The compressed vector from ScatNet with PCA |
| $\mathcal{Z}_\alpha$ | The compressed tensor from FrScatNet with FMF and with fractional order $\alpha$ |
| $\psi$ | The complex band-pass filter |
| $U[p]$ | The scattering propagator with frequency-decreasing $p$-th path |
| $S[p]$ | The scattering operator |
| $U_\alpha[p]$ | The fractional scattering propagator with frequency-decreasing $p$-th path |
| $\Theta_\alpha$ | The fractional convolution |
| $S_\alpha[p]$ | The fractional scattering operator |

## 2. Generative Scattering Networks (GSNs)

In this section, we first briefly introduce the Generative Scattering Networks (GSNs) [66], whose structure is shown in Fig. 1. The input $M$th-order tensor $\mathcal{X} \in \mathbb{R}^{N_1 \times N_2 \times \cdots \times N_K}$, where $\mathbb{R}$ denotes real domain and each $N_i$, $i=1,2,\ldots,K$, addresses the $i$-mode of $\mathcal{X}$, is first feed into the Wavelet Scattering Networks (ScatNets) to obtain the ScatNet features (or ScatNet embeddings) $\mathcal{Y} \in \mathbb{R}^{M_1 \times M_2 \times \cdots \times M_L}$, whose dimensions are then reduced by PCA to obtain an implicit vector $\mathbf{z} \in \mathbb{R}^U$, which is then feed into the generator $G_1$ to obtain the generated output tensor $\tilde{\mathcal{X}} \in \mathbb{R}^{N_1 \times N_2 \times \cdots \times N_K}$. That is, the generative network $G_1$ is seen as the inverse problem of ScatNets.

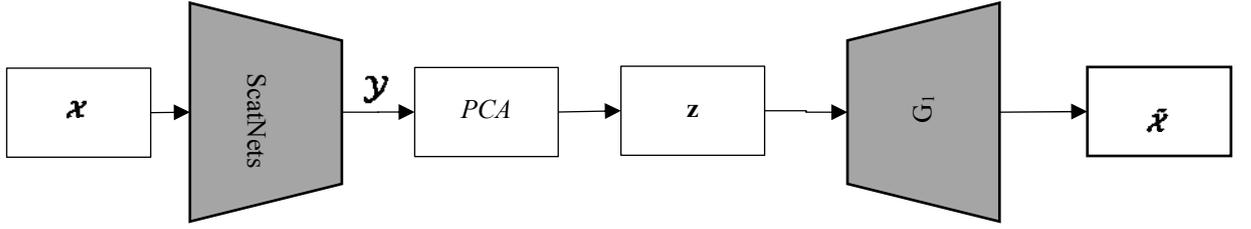

Fig. 1 The structure of GSNs with PCA dimensional reduction method. The input $\mathcal{X}$ enters the ScatNets to obtain the ScatNet features $\mathcal{Y}$, which are then compressed by PCA to obtain an implicit vector **z**, then the output $\tilde{\mathcal{X}}$ is obtained from the generator $G_1$.

The main components of GSNs include ScatNets, PCA dimensional reduction method and the generative network $G_1$. These components of GSNs are introduced as follows.

### 2.1 Wavelet Scattering Networks (ScatNets)

In this section, the wavelet scattering networks [67, 68] will be introduced.

Let the complex band-pass filter $\psi_\lambda$ be constructed by scaling and rotating a filter $\psi$ respectively by $2^j$ and $\delta$, that is,

$$\psi_\lambda(t) = 2^{2j}\psi(2^j \delta^{-1} t), \quad \lambda = 2^j \delta \tag{1}$$

with $0 \leq j \leq J-1$, and $\delta = k\pi/K$, $k = 0,1,\ldots,K-1$.

The wavelet-modulus coefficients of $x$ are given by

$$U[\lambda]x = |x * \psi_\lambda| \tag{2}$$

The scattering propagator $U[p]$ is defined by cascading wavelet-modulus operators [67, 68]

$$U[p]x = U[\lambda_m]\cdots U[\lambda_2]U[\lambda_1]x = \left|\left||x * \psi_{\lambda_1}| * \psi_{\lambda_2}\right|\cdots * \psi_{\lambda_m}\right|, \tag{3}$$

where $p = (\lambda_1, \lambda_2, ..., \lambda_m)$ are the frequency-decreasing paths, that is, $|\lambda_k| \geq |\lambda_{k+1}|, k=1,2,...,m-1$. Note that $U[\varnothing]x = x$, and $\varnothing$ express empty set.

Scattering operator $S_J$ performs a spatial averaging on a domain whose width is proportional to $2^J$:

$$S[p]x = U[p]x * \phi_J = U[\lambda_m]\cdots U[\lambda_2]U[\lambda_1]x * \phi_J = \left\lVert \left\lvert x * \psi_{\lambda_1} \right\rvert * \psi_{\lambda_2} \right\rvert \cdots * \psi_{\lambda_m} \right\rVert * \phi_J. \qquad (4)$$

The wavelet scattering network is shown in Fig. 2. The network nodes of the layer $m$ correspond to the set $P^m$ of all paths $p = (\lambda_1, \lambda_2, ..., \lambda_m)$ of length $m$. This $m$-th layer stores the propagated signals $\{U[p]x\}_{p \in P^m}$ and outputs the scattering coefficients $\{S[p]x\}_{p \in P^m}$. The output is obtained by cascading the scattering coefficients of every layers.

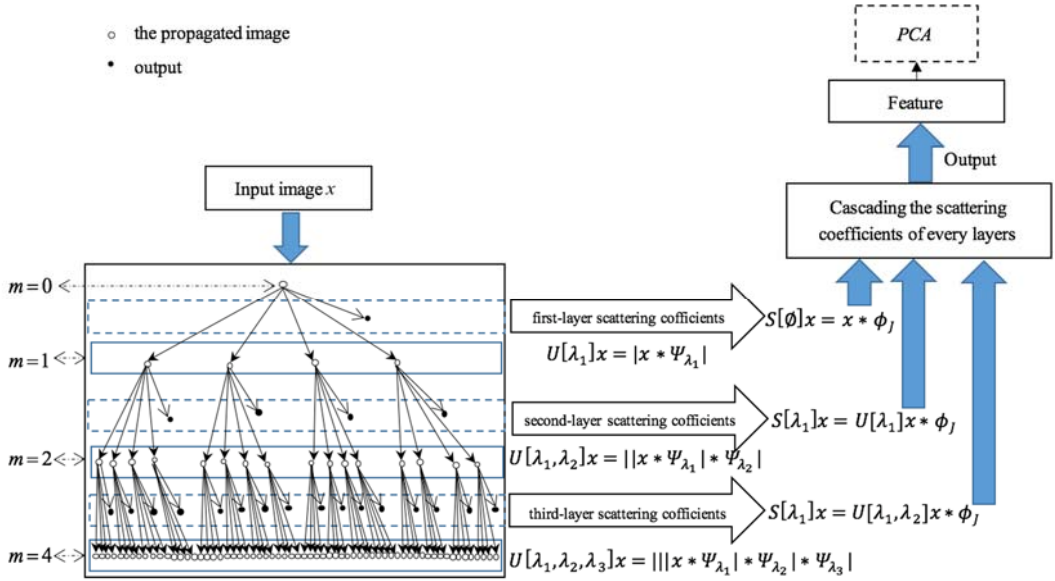

Fig. 2 Three-layer wavelet scattering network. "○" denotes applying scattering propagator $U$ to obtain the intermediate results. "●" denotes the outputs of the scattering network, that is, scattering coefficients, which are cascaded layer by layer to obtain the ScatNet features (or ScatNet embeddings).

Note that $x$ in (2) can be one-dimensional data $\mathbf{x} \in \mathbb{R}^{N_1}$, two-dimensional data $\mathbf{X} \in \mathbb{R}^{N_1 \times N_2}$, and also third-order tensor $\mathcal{X} \in \mathbb{R}^{N_1 \times N_2 \times N_3}$, which can be seen as $N_3$ two-dimensional data $\mathbf{X} \in \mathbb{R}^{N_1 \times N_2}$ and the ScatNet deals with these $\mathbf{X}$ one by one and then superimpose these results as output features. According to [67], if we feed the input $\mathcal{X} \in \mathbb{R}^{N_1 \times N_2 \times N_3}$ into the ScatNet, then we can obtain the ScatNet features (or ScatNet embeddings) as follows

$$\mathcal{Y} = S[p]\mathcal{X} \in \mathbb{R}^{N_3 \times (1+LJ+L^2J(J-1)/2) \times (N_1/2^J) \times (N_2/2^J)} \qquad (5)$$

where $N_3$ is the number of input sample channels, $N_1$ and $N_2$ are the width and the height of input sample, respectively. $N_1/2^J$ and $N_2/2^J$ are the width and the height of output features. $J$ is a scale factor and $L$ is the number of rotation angle. Note that the number of feature map in the first layer, in the second layer, and in the third layer are 1, $LJ$, and $L^2J(J-1)/2$, respectively.

## 2.2 Feature dimensional reduction by PCA

Principal Component Analysis [69] (PCA) is a simple but commonly used linear dimensional reduction method, which generally shows good results on one-dimensional signal compression. In Generative Scattering Networks (GSNs) [66], PCA is used to reduce the dimension of ScatNets features and then obtain the implicit vector **z**, which is then input the following generator $G_1$.

Therefore, an input tensor $\mathcal{X} \in \mathbb{R}^{N_1 \times N_2 \times N_3}$ feed into the ScatNets to obtain ScatNet features $\mathcal{Y}$ in (5). Then, the dimension of $\mathcal{Y}$ is reduced by PCA to obtain an implicit vector

$$\mathbf{z} = \text{PCA}(\mathcal{Y}) \tag{6}$$

where the PCA(.) mean PCA algorithms, respectively.

The obtained implicit vector **z** is then input the generator networks $G_1$ to get the generated image in the following Section 2.3.

## 2.3 Generator Networks in GSNs

The generative network $G_1$ of GSNs is a neural network, which is similar to the generator of DCGAN [22], which inverts the whitened scattering embedding on training samples. The structure of generator network $G_1$ is illustrated in Fig. 3, which includes fully connection layer (FC), batch normalization layer [73], bilinear upsamling (Upsample) layer and convolutional layer (Conv2d) [74-76] with kernel size 7×7. Except the last layer uses tanh activation function, other layers use default Relu [77] activation function.

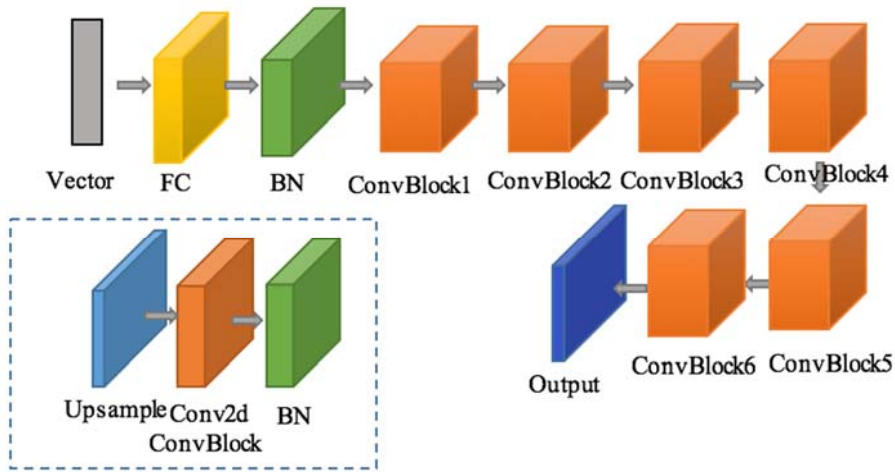

Fig. 3 The structure of generative network $G_1$ in GSNs. Note that the implicit vector **z** is the input of the generative

network G₁.

GSNs with PCA as dimensional reductional method choose L₁-norm loss function, and solve the following optimization problem [78]:

$$g_1 = \min\ Loss_{L_1}(\mathcal{X}, \tilde{\mathcal{X}}) = \min\ \frac{1}{N}\sum_{i=1}^{N}\left|\mathcal{X}^{(i)} - \tilde{\mathcal{X}}^{(i)}\right| \tag{7}$$

where $\mathcal{X}$ represents the input data, and $\tilde{\mathcal{X}}$ represents the generative data, $\mathcal{X}^{(i)}$ represents the *i*-th input sample, $\tilde{\mathcal{X}}^{(i)}$ represents the *i*-th generative sample, and

$$\tilde{\mathcal{X}} = G_1\left(\mathrm{PCA}\left(S[p]\mathcal{X}\right)\right) \tag{8}$$

where $S[p]\mathcal{X}$ denotes the feature extraction process with ScatNets, PCA(.) represents the feature dimensional reduction method is PCA. $G_1(.)$ represents the generative network G₁ shown in Fig. 3. The optimization problems in (7) is then solved by Adam optimizer [79] using the default hyperparameters.

## 3. Generative Fractional Scattering Networks (GFRSNs)

In this section, we introduce the proposed Generative Fractional Scattering Networks (GFRSNs), whose structure is shown in Fig. 4. The input $\mathcal{X} \in \mathbb{R}^{N_1 \times N_2 \times \cdots \times N_K}$ is first feed into the Fractional wavelet Scattering Networks (FrScatNets) to obtain the FrScatNet features (or FrScatNet embeddings) $\mathcal{Y}_\alpha \in \mathbb{R}^{M_1 \times M_2 \times \cdots \times M_L}$, whose dimensions are then reduced by the proposed Feature-Map Fusion (FMF) method to obtain an implicit tensor $\mathcal{Z}_\alpha \in \mathbb{R}^{O_1 \times O_2 \times \cdots \times O_K}$, which are then feed into the generator G₂ to obtain the generated output tensor $\tilde{\mathcal{X}}_\alpha \in \mathbb{R}^{N_1 \times N_2 \times \cdots \times N_K}$. That is, the generative network G₂ is seen as the inverse problem of FrScatNets.

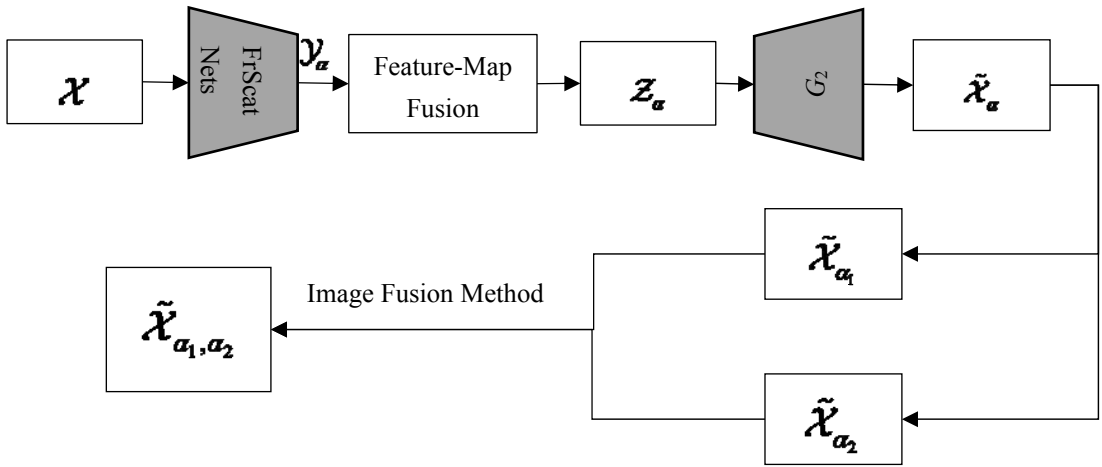

Fig. 4 The structure of Generative Fractional Scattering Networks (GFRSNs). The input $\mathcal{X}$ enters the FrScatNets to obtain the FrScatNet features $\mathcal{Y}_\alpha$, which are then compressed by Feature-Map Fusion method to obtain an implicit tensor $\mathcal{Z}_\alpha$, then the output $\tilde{\mathcal{X}}_\alpha$ is obtained from generator G₂.

The final output of GFRSNs $\tilde{x}_{\alpha_1,\alpha_2}$ is obtained by the image fusion of two generative images of different fractional orders $\alpha_1$ and $\alpha_2$.

The main components of GFRSNs include FrScatNets, Feature-Map Fusion dimensional reduction method and the generative network $G_2$. These components of GFRSNs are introduced as follows.

## 3.1 Fractional wavelet Scattering Networks (FrScatNets)

In this subsection, the Fractional wavelet Scattering Networks (FrScatNets) [70] will be briefly introduced.

Similar to (2), the fractional wavelet modulus coefficients of $x$ are given by

$$U_\alpha[\lambda]x = |x \Theta_\alpha \psi_\lambda| \tag{9}$$

where $\Theta_\alpha$ is the fractional convolution defined by [80]

$$x(t)\Theta_\alpha \psi_\lambda(t) = e^{-\frac{j}{2}t^2 \cot\theta} \left[ (x(t)e^{\frac{j}{2}t^2 \cot\theta}) * \psi_\lambda(t) \right] \tag{10}$$

where the parameter $\alpha$ is the fractional order and $\theta = \alpha\pi/2$ represents the rotation angle. When $\alpha = 1$, the fractional convolution in (10) reduces to conventional convolution in (2).

The fractional scattering propagator $U_\alpha[p]$ is defined by cascading fractional wavelet modulus operators [70]

$$U_\alpha[p]x = U_\alpha[\lambda_m]\cdots U_\alpha[\lambda_2]U_\alpha[\lambda_1]x = \left\| |x\Theta_\alpha \psi_{\lambda_1}| \Theta_\alpha \psi_{\lambda_2} \right| \cdots \Theta_\alpha \psi_{\lambda_m} \right|, \tag{11}$$

where $p = (\lambda_1, \lambda_2, ..., \lambda_m)$ are the frequency-decreasing paths, that is, $|\lambda_k| \geq |\lambda_{k+1}|, k = 1,2,...,m-1$. Note that $U_\alpha[\varnothing]x = x$, and $\varnothing$ express empty set.

The fractional scattering operator $S_\alpha$ performs a spatial averaging on a domain whose width is proportional to $2^J$:

$$S_\alpha[p]x = U_\alpha[p]x\Theta_\alpha\phi_J = U_\alpha[\lambda_m]\cdots U_\alpha[\lambda_2]U_\alpha[\lambda_1]x\Theta_\alpha\phi_J = \left\| |x\Theta_\alpha\psi_{\lambda_1}|\Theta_\alpha\psi_{\lambda_2}\right|\cdots\Theta_\alpha\psi_{\lambda_m}\right|\Theta_\alpha\phi_J. \tag{12}$$

The FrScatNets is shown in Fig. 5. The network nodes of the layer $m$ correspond to the set $P^m$ of all paths $p = (\lambda_1, \lambda_2, ..., \lambda_m)$ of length $m$. This $m$-th layer stores the propagated signals $\{U_\alpha[p]x\}_{p \in P^m}$ and outputs the fractional scattering coefficients $\{S_\alpha[p]x\}_{p \in P^m}$. The output is obtained by cascading the fractional scattering coefficients of every layers. Note that when $\alpha = 1$, the FrScatNets in (12) defaults to conventional ScatNets in (4) since the fractional convolution in (10) reduces to conventional convolution in (2).

Note that FrScatNets keep the advantages of ScatNets, for example, without learning, translation invariant property, linearize deformations, a bit of parameters, etc. Compared to ScatNets, FrScatNets adds a free parameter $\alpha$ which represents fractional order. With $\alpha$ continuous growth from 0 to 2, FrScatNets can show the characteristics of an image from time domain to frequency domain. Thus, FrScatNets provide more fractional domain choices for the feature extraction of input data. Furthermore, for the image generated task in this paper, we may obtain many generated images from FrScatNets embeddings of different fractional orders $\alpha_i$ and then fuse these generated images to further improve the quality of generated image.

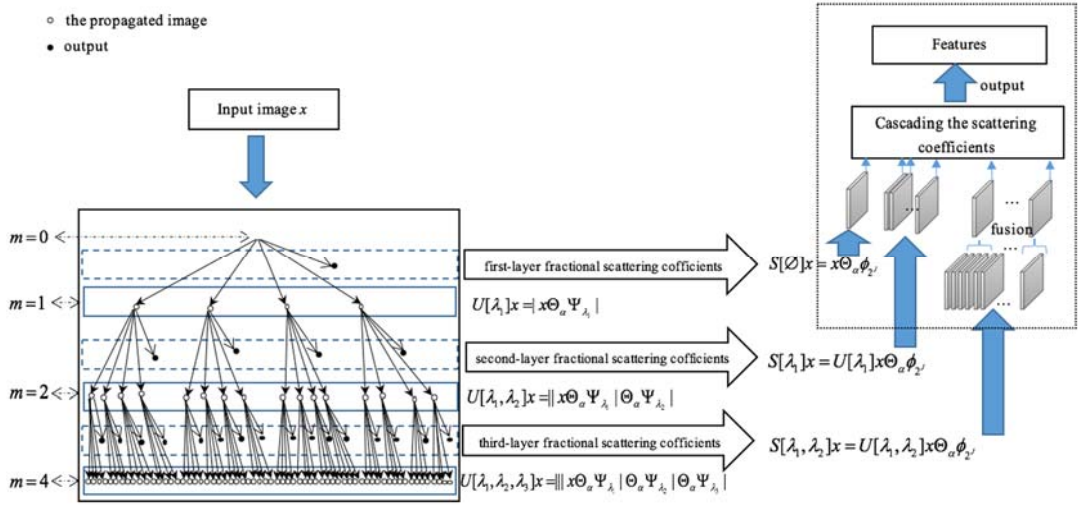

Fig. 5 The fractional wavelet scattering network and the Feature-Map Fusion dimensional reduction method.

If we feed the input $\mathcal{X} \in \mathbb{R}^{N_1 \times N_2 \times N_3}$ into the FrScatNet, then we can obtain the FrScatNet features (or FrScatNet embeddings)

$$\mathcal{Y}_\alpha = S_\alpha[p]\mathcal{X} \in \mathbb{R}^{N_3 \times (1+LJ+L^2 J(J-1)/2) \times (N_1/2^J) \times (N_2/2^J)} \quad (13)$$

Note that the size of output features of FrScatNets is the same as ScatNets, whose size is shown in (5).

## 3.2 Feature dimensional reduction by Feature-Map Fusion (FMF) method

In this subsection, we reduce the dimension of the FrScatNet features by the proposed FMF method.

Why we propose the FMF method instead of using the mature PCA [69] algorithm? Because we find that FrScatNets has a hierarchical tree structure, that is, the features of different layers from FrScatNets have hierarchical information, PCA algorithm does not consider the differences of semantic information contained in the feature of FrScatNets with different layers. Therefore, we need a dimensional reduction method that can consider the hierarchical information of FrScatNets. Since the number of feature map in the first layer, in the second layer, and in the third layer are 1, $LJ$, and $L^2J(J-1)/2$, respectively. Obviously, the third layer has the largest number of feature maps. Therefore, we only fuse the feature maps from the third layer of FrScatNets to significantly reduce the data dimension. The fusion method is very simple, that is, we obtain a new feature map by simply taking the average of every $L(J-1)/2$ feature maps, which obtain $LJ$ new feature maps after applying the FMF method on the output of the third layer of FrScatNets. The dotted box of Fig. 5 illustrates the proposed FMF method.

Therefore, an input tensor $\mathcal{X} \in \mathbb{R}^{N_1 \times N_2 \times N_3}$ feed into the FrScatNets to obtain FrScatNet features $\mathcal{Y}_\alpha$ in (13), which are then processed by the FMF method, obtaining an implicit tensor

$$\mathcal{Z}_\alpha = \text{FMF}(\mathcal{Y}_\alpha) \in \mathbb{R}^{N_3 \times (1+LJ+LJ) \times (N_1/2^J) \times (N_2/2^J)} \tag{14}$$

whose size is significantly smaller than the size shown in (13) without using the FMF method. Note that FMF(.) means performing the Feature-Map Fusion method.

The obtained implicit tensor $\mathcal{Z}_\alpha$ is then input the generator networks $G_2$ to get the generated image in the following subsection.

## 3.3 Generative Networks in GFRSNs

The generative network $G_2$ of GFRSNs is also a deconvolutional neural networks, which has a similar generator as DCGAN [22], which inverts the fractional scattering embeddings on training samples. The structure of generative network $G_2$ is illustrated in Fig. 6, which includes fully convolutional layer (Fully Conv) [81], and several convolution blocks which consist of bilinear upsampling (UP), two convolutional layers (Conv) with 3×3 kernel size, batch normalization [73], Relu [77] (the activation function of the last convolution layer is tanh). For the CelebA dataset [82], we use the whole structure shown in Fig. 6, while for the CIFAR-10 dataset [83], ConvBlock4 is removed.

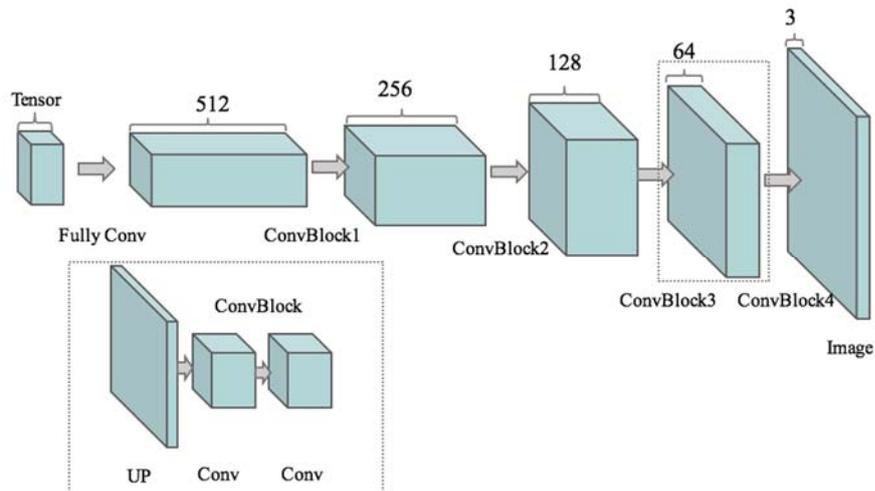

Fig. 6 The architecture of generative network $G_2$ in GFRSNs. Note that the implicit tensor $\mathcal{Z}_\alpha$ in (14) is the input of the generative network $G_2$.

GFRSNs also chooses $L_1$-norm loss function, and solves the following optimization problem:

$$g_2 = \min \; Loss_{L_1}\left(\mathcal{X}, \tilde{\mathcal{X}}_\alpha\right) = \min \; \frac{1}{N}\sum_{i=1}^{N}\left|\mathcal{X}^{(i)} - \tilde{\mathcal{X}}_\alpha^{(i)}\right| \tag{15}$$

where $\tilde{\mathcal{X}}_\alpha$ represents the generative data, $\tilde{\mathcal{X}}_\alpha^{(i)}$ represents the $i$-th generative sample, and

$$\tilde{\mathcal{X}}_\alpha = G_2\left(\text{FMF}\left(S_\alpha[p]\mathcal{X}\right)\right) \tag{16}$$

where $S_\alpha[p]\mathcal{X}$ denotes the feature extraction process with FrScatNets, FMF(.) represents that the feature dimensional reduction method is Feature-Map Fusion. $G_2$(.) represents the generative network $G_2$ shown in Fig. 6.

The optimization problem in (15) is then solved by Adam optimizer [79] using the default hyperparameters.

## 3.4 Image Fusion

Generative Scattering Networks (GSNs) [66] embeds the input using ScatNets [67, 68], which obtain only one embedding. In contrast to GSNs, the proposed Generative Fractional Scattering Networks (GFRSNs) embeds the input using FrScatNets [70], which can obtain many embeddings since FrScatNets has an additional fractional order $\alpha$, therefore, we can embed the input in different fractional order domains. These FrScatNets embeddings may extract many different but complementary features from the input. Can we effectively use these embeddings? We can use these FrScatNets embeddings to generate many images, and further improve the quality of synthesized images by using the image fusion methods, which are shown in Fig. 4. In this paper, we use a simple image fusion method which is shown as follows

$$\tilde{\mathcal{X}}_{\alpha_1,\alpha_2} = \lambda\tilde{\mathcal{X}}_{\alpha_1} + (1-\lambda)\tilde{\mathcal{X}}_{\alpha_2} \tag{17}$$

where $\lambda$ is the balanced parameter and is set to 0.5 in this paper. That is, we use the simple average method for image fusion, which is shown in Fig. 7.

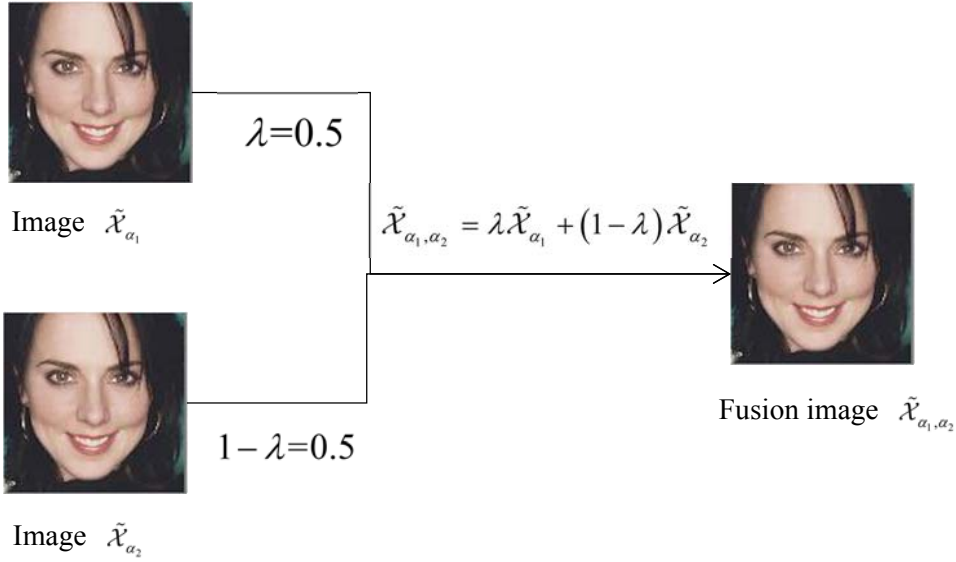

Fig. 7 Image fusion with average method.

## 4. Numerical Experiments

In this section, we evaluate the quality of the generative images by the proposed GFRSNs in Fig. 4 with several experiments. The quality of the generative images is evaluated with two criteria: Peak Signal to Noise Ratio (PSNR) [84] and Structural Similarity (SSIM) [85].

We performed the experiments on two datasets that have different levels of variabilities: CIFAR-10 and CelebA. CIFAR-10 dataset includes 50000 training images and 10000 testing images, whose sizes are 32×32×3. In all experiments on CIFAR-10 dataset after image grayscale preprocessing, the number of rotation angle $L$ is set to 8 and the fractional scattering averaging scale is set to $2^J = 2^3 = 8$ which means we linearize translations and deformations of up to 8 pixels, so, the size of the output features from FrScatNets according to Eq. (13) is 1×217×4×4, which is then, after FMF method according to Eq. (14), reduced to 1×49×4×4 (the size of implicit tensor $\mathcal{Z}_\alpha$). In addition, CelebA dataset contains thousands of images, we choose 65536 training images and 16384 test images, whose sizes are 128×128×3. In all experiments on CelebA dataset after image grayscale preprocessing, the number of rotation angle $L$ is set to 8 and the fractional scattering averaging scale is set to $2^J = 2^4 = 16$ which means we linearize translations and deformations of up to 16 pixels, so, the size of the output features from FrScatNets according to Eq. (13) is 1×417×8×8, which is then, after FMF method according to Eq. (14), reduced to 1×65×8×8 (the size of implicit tensor $\mathcal{Z}_\alpha$).

Table 2 shows the core parameters of FrScatNet and their settings on CIFAR-10 dataset and CelebA dataset.

Table 2 The core parameters of FrScatNet with and without feature dimensional reduction.

| Parameters | Descriptions | Settings | |
|---|---|---|---|
| | | CIFAR-10 | CelebA |
| $N_1 \times N_2 \times N_3$ | The size of input image | 32×32×1 | 128×128×1 |
| $J$ | The fractional scattering averaging scale | 3 | 4 |
| $L$ | The number of rotation angle | 8 | 8 |
| $\alpha$ | The fractional order. When $\alpha =1$, FrScatNets defaults to ScatNets. | $0 \leq \alpha \leq 2$ | $0 \leq \alpha \leq 2$ |
| $N_3 \times (1 + LJ + \frac{L^2 J(J-1)}{2}) \times \frac{N_1}{2^J} \times \frac{N_2}{2^J}$ | The size of FrScatNets features $y_\alpha$, which is the output features of FrScatNets without feature dimensional reduction. | 1×217×4×4 | 1×417×8×8 |
| $N_3 \times (1 + 2 \times LJ) \times \frac{N_1}{2^J} \times \frac{N_2}{2^J}$ | The size of implicit tensor $z_\alpha$ which is obtained by applying the FMF method to the out features of FrScatNets. | 1×49×4×4 | 1×65×8×8 |

In the following, we first compare the visual quality of generated images with different feature dimensionality reduction methods in the framework of GFRSNs. Then, we compare the visual quality of generated images with FrScatNets. Finally, we compare the visual quality of fused images and un-fused images. The following experiments are implemented using PyTorch on a PC machine, which sets up Ubuntu 16.04 operating system and has an Intel(R) Core(TM) i7-8700K CPU with speed of 3.70 GHz and 32 GB RAM, and has also two NVIDIA GeForce GTX1080-Ti GPUs.

## 4.1 Image generative results with different dimensionality reduction methods

In this subsection, we compare the results of generative image quality with two different dimensionality reduction methods: PCA method [69] and the proposed Feature-Map Fusion method. We set the fractional orders $\alpha_1=\alpha_2=1$, that is, we use the conventional ScatNets to extract the features from input $\mathcal{X}$ for simplicity.

For the PCA based GFRSNs, the flow chart is shown in Fig. 1. For CIFAR-10 dataset, the size of implicit vector **z** is 49×4×4=784, and for CelebA dataset, the size of implicit vector **z** is 65×8×8=4160. We use the PyTorch code of generative scattering networks[1], which is provided by Tomás Angles. The training and testing curves on CIFAR-10 and CelebA datasets are shown in Fig. 8. The PSNR and SSIM on CIFAR-10 and CelebA datasets are shown in the second columns of the Table 3 and the Table 4, respectively. As can be seen from the Fig. 8 and two Tables, the scores of PSNR in training dataset (Train PSNR) and the scores of SSIM in training dataset (Train SSIM) are very good for the PCA based GFRSNs, however, the scores of

---
[1] https://github.com/tomas-angles/generative-scattering-networks

PSNR in testing dataset (Test PSNR) and the scores of SSIM in testing dataset (Test SSIM) are slight blur. This phenomenon indicates that overfitting problem has occurred by using the PCA based GFRSNs. The overfitting problem is also validated by the visual results of generated image shown in the Fig. 10 and the Fig. 10 . We guess the reason maybe that the output feature of FrScatNets $\mathcal{Y}_\alpha$ in (16) is a 4th-order tensor, which is performed by PCA to obtain an implicit vector **z**, this process loses the correlations between various dimensions of the data. Therefore, we consider to use FMF as the dimensionality reduction method to better maintain the structures of the input data.

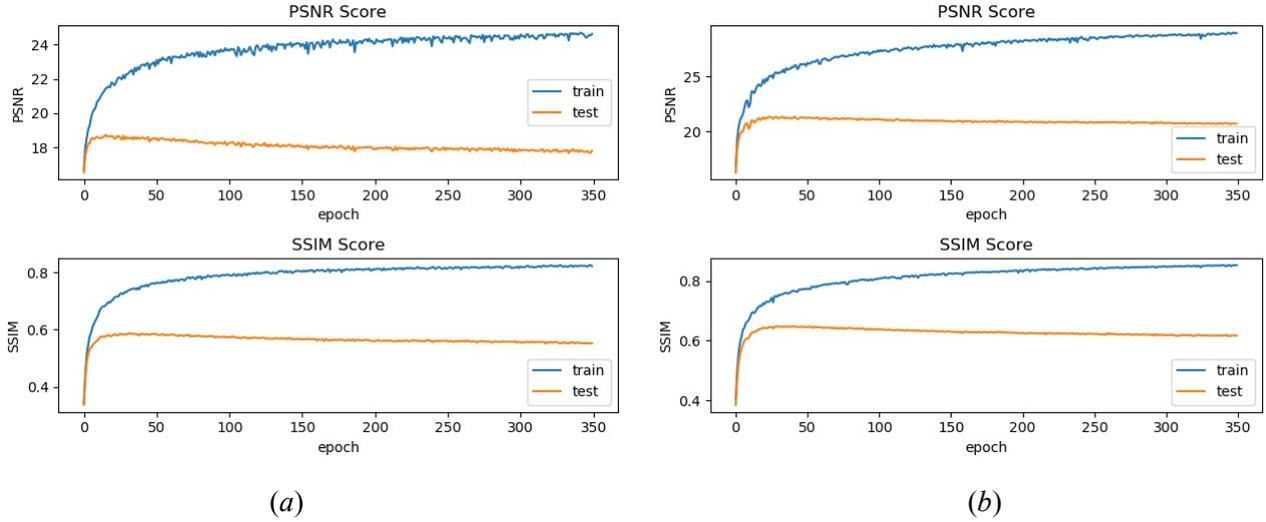

(*a*)            (*b*)

Fig. 8 The training and testing curves on CIFAR-10 and CelebA datasets by using PCA dimensionality reduction method in GFRSNs with fractional orders $\alpha_1 = \alpha_2 = 1$. (*a*) The results of Loss, PSNR and SSIM on CIFAR-10 dataset; (*b*) The results of Loss, PSNR and SSIM on CelebA dataset.

Table 3 PSNR and SSIM scores of training and testing images from FrScatNets with fractional orders $\alpha_1 = \alpha_2 = 1$ on CIFAR-10 datasets. Increased means the percentages of relative improvements of FMF over PCA.

|  | PCA | Feature-Map Fusion | Increased (%) |
|---|---|---|---|
| Train PSNR | **24.61425** | 20.88963 | -15.1 |
| Test PSNR | 17.97365 | **19.24235** | 6.6 |
| Train SSIM | **0.82303** | 0.70257 | -14.6 |
| Test SSIM | 0.55516 | **0.63264** | 12.2 |

Table 4 PSNR and SSIM scores of training and testing images from FrScatNets with fractional orders $\alpha_1 = \alpha_2 = 1$ on CelebA datasets. Increased means the percentages of relative improvements of FMF over PCA.

|  | PCA | Feature-Map Fusion | Increased (%) |
|---|---|---|---|
| Train PSNR | **28.20955** | 24.02802 | -14.8 |
| Test PSNR | 20.69532 | **21.99852** | 6.0 |
| Train SSIM | **0.84082** | 0.73765 | -12.3 |
| Test SSIM | 0.62062 | **0.70366** | 11.8 |

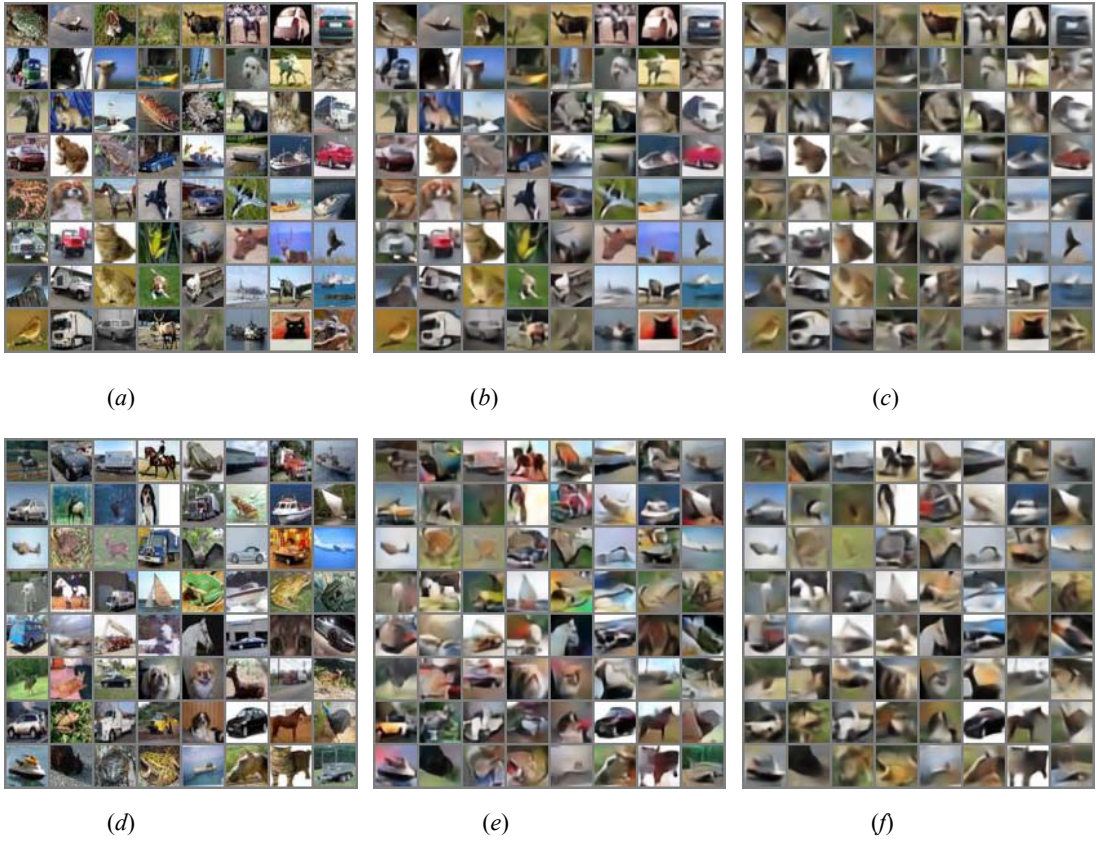

Fig. 9 Generated images with different dimensionality reduction methods on CIFAR-10 dataset. (*a*) Original training images; (*b*)Generative training images with PCA based GFRSNs; (*c*) Generative training images with FMF based GFRSNs; (*d*) Original testing images; (*e*)Generative testing images with PCA based GFRSNs; (*f*) Generative testing images with FMF based GFRSNs.

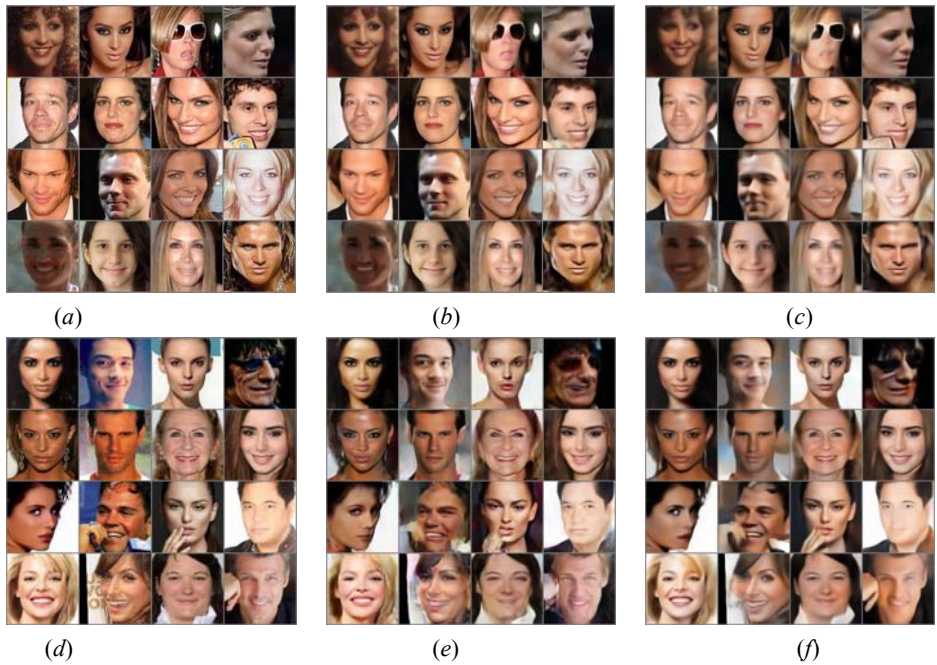

Fig. 10 Generated images with different dimensionality reduction methods on CelebA dataset. (*a*) Original training images; (*b*)Generative training images with PCA based GFRSNs; (c) Generative training images with FMF based GFRSNs; (*d*) Original testing images; (*e*)Generative testing images with PCA based GFRSNs; (*f*) Generative testing images with FMF based GFRSNs.

For the proposed FMF based GFRSNs, the flow chart is shown in Fig. 4. The size of implicit tensor $\mathcal{Z}_{\alpha_i}$ is 1×49×4×4 on CIFAR-10, and for CelebA dataset, the size of implicit tensor $\mathcal{Z}_{\alpha_i}$ is 1×65×8×8. The training and testing curves on CIFAR-10 and CelebA datasets are shown in Fig. 11. The PSNR and SSIM on CIFAR-10 and CelebA datasets are shown in the third column of the Table 3 and Table 4, respectively. As can be seen from the Fig. 11 and two Tables, Train PSNR and Train SSIM of the FMF based GFRSNs are slightly worse than the PCA based GFRSNs on CIFAR-10 dataset and on CelebA dataset, however, Test PSNR and Test SSIM of the proposed FMF based GFRSNs are better than PCA based GFRSNs. For example, Test PSNR and Test SSIM are relatively increased by 6.6% and 12.2% when compared to PCA based GFRSNs on CIFAR-10 datasets. Test PSNR and Test SSIM are relatively increased by 6.0% and 11.8% when compared to PCA based GFRSNs on CelebA dataset. The experimental results show that overfitting problem on testing dataset is effectively alleviated with the FMF dimensionality reduction method. The conclusion is also validated by the visual results of generated image shown in the Fig. 9 (*c*), (*f*) and Fig. 10 (*c*), (*f*).

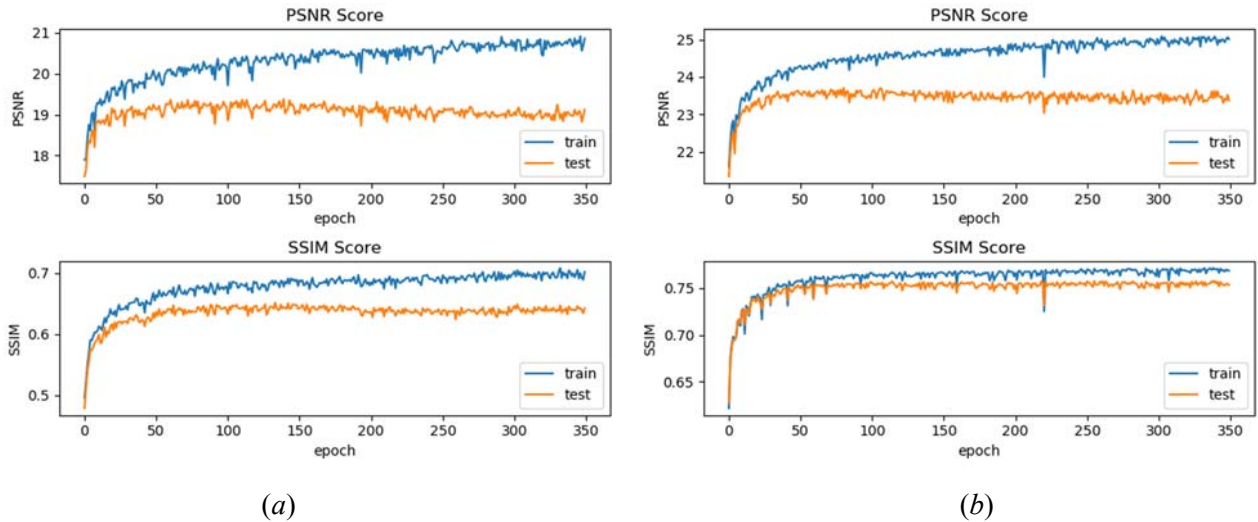

(*a*)  (*b*)

Fig. 11 The training and testing curves on CIFAR-10 and CelebA datasets by using FMF dimensionality reduction method in GFRSNs with fractional orders $\alpha_1 = \alpha_2 = 1$. (*a*) The results of PSNR and SSIM on CIFAR-10 dataset; (*b*) The results of PSNR and SSIM on CelebA dataset.

Since the proposed FMF method performs better than PCA on testing datasets, that is, FMF has better generalization performance in the framework of GFRSNs, therefore, we use FMF method in the following two experiments.

## 4.2 Image generative results with different fractional order α

In this subsection, we explore the impact of fractional order $\alpha$ on the quality of generated image using the framework of GFRSNs shown in Fig. 5. The other parameter settings of FrScatNet are shown in Table 2. We choose the $L_1$ loss function in (15) and train the generator by Adam optimizer using the default hyperparameters.

In this subsection, we use two-dimensional fractional Morlet wavelet to construct the FrScatNets. For the two-dimensional fractional wavelet, two fractional orders $\alpha_1$ and $\alpha_2$ are needed to determine the rotational angle. The angle is defined as $\theta = \alpha\pi/2$ ranging from 0 to $\pi$, so the fractional orders $\alpha_1$ and $\alpha_2$ change from 0 to 2. To save computation time, we fix one order as 1.00 and the other one changes within the range zero to two for computing the fractional scattering coefficients. The chosen values are 0.05, 0.10, 0.40, 0.70, 1.00, 1.30, 1.60, 1.90, and 1.95, respectively. These above parameter settings are the same as [70]. Note that FrScatNets reduces to conventional ScatNets when $\alpha_1 = \alpha_2 = 1.00$.

The PSNR and SSIM of the generated images from FrScatNets on CIFAR-10 dataset and on CelebA dataset are shown in Table 5 and Table 6. From Table 5, we can see that for CIFAR-10 datasets, the best results are generally not appeared by using FrScatNets with $(\alpha_1, \alpha_2)=(1.00, 1.00)$, which means that FrScatNets with some fractional order choice of $(\alpha_1, \alpha_2)$ obtain better embeddings than the conventional ScatNets. For example, both the PSNR and the SSIM results are very good by using the FrScatNets with $(\alpha_1, \alpha_2)=(0.40, 1.00)$, whose Train PSNR, Test PSNR, Train SSIM and Test SSIM are increased respectively by 1.4%, 1.4%, 1.4% and 3.9% when compared to those of ScatNets. From Table 6, we can see that for CelebA dataset, both the PSNR and the SSIM results are also very good by using the FrScatNets with $(\alpha_1, \alpha_2)=(1.60, 1.00)$, whose Train PSNR, Test PSNR, Train SSIM and Test SSIM are increased respectively by 0.8%, 1.4%, 2.4% and 3.8% when compared to those of ScatNets.

Table 5 The results with FrScatNets on CIFAR-10 dataset. The best results are shown in bold. In the third column, "No" means un-fused image and "Yes" means fused image. Increased1, Increased2, Increased3, Increased4 mean the percentages of relative improvements on Train PSNR, Test PSNR, Train SSIM, Test SSIM of FrScatNets of various fractional orders $(\alpha_1, \alpha_2)$ over conventional ScatNets, respectively.

| Index | $(\alpha_1, \alpha_2)$ | Fusion or not? | Train PSNR | Increased1(%) | Test PSNR | Increased2(%) | Train SSIM | Increased3(%) | Test SSIM | Increased4(%) |
|---|---|---|---|---|---|---|---|---|---|---|
| 1 | (1.00,1.00) | No | 20.88963 | 0 | 19.24235 | 0 | 0.70257 | 0 | 0.63264 | 0 |
| 2 | (0.05,1.00) | No | 19.23995 | -7.9 | 16.8271 | -12.6 | 0.57644 | -18 | 0.4681 | -26 |
|   |   | Yes | 20.57486 | -1.5 | 18.99788 | -1.3 | 0.67675 | -3.7 | 0.60127 | -5 |
| 3 | (0.10,1.00) | No | 17.77867 | -14.8 | 15.00028 | -22 | 0.44948 | -36 | 0.28296 | -55.3 |
|   |   | Yes | 20.23647 | -3.1 | 18.03799 | -6.3 | 0.63035 | -10.3 | 0.52798 | -16.5 |
| 4 | (0.40,1.00) | No | 21.18276 | 1.4 | **19.52272** | 1.4 | 0.71221 | 1.4 | **0.65852** | 3.9 |

| Index | ($α_1$, $α_2$) | Fusion or not? | Train PSNR | Increased 1 (%) | Test PSNR | Increased 2 (%) | Train SSIM | Increased3 (%) | Test SSIM | Increased4 (%) |
|---|---|---|---|---|---|---|---|---|---|---|
| | | Yes | 21.89039 | 4.6 | **20.23685** | 4.9 | 0.73349 | 4.2 | 0.68317 | 7.4 |
| 5 | (0.70,1.00) | No | 20.82666 | -0.3 | 19.02157 | -1.1 | 0.69564 | -1 | 0.63576 | 0.5 |
| | | Yes | 21.77986 | 4.1 | 20.06666 | 4.1 | 0.72841 | 3.5 | 0.67598 | 6.4 |
| 6 | (1.30,1.00) | No | 20.84246 | -0.2 | 18.9987 | -1.3 | 0.6951 | -1.1 | 0.63441 | 0.3 |
| | | Yes | 21.75738 | 4.1 | 20.03336 | 3.9 | 0.72707 | 3.4 | 0.6742 | 6.2 |
| 7 | (1.60,1.00) | No | **21.22009** | 1.6 | 19.37945 | 0.7 | **0.7151** | 1.8 | 0.65834 | 3.9 |
| | | Yes | **21.94697** | 4.8 | 20.20283 | 4.8 | **0.73605** | 4.5 | 0.68451 | 7.6 |
| 8 | (1.90,1.00) | No | 18.02042 | -13.7 | 15.0661 | -21.7 | 0.46772 | -33.4 | 0.29763 | -53 |
| | | Yes | 20.36688 | -2.5 | 18.07771 | -6.1 | 0.63702 | -9.3 | 0.53307 | -15.7 |
| 9 | (1.95,1.00) | No | 19.13651 | -8.4 | 16.46899 | -14.4 | 0.56153 | -20.1 | 0.43501 | -31.2 |
| | | Yes | 20.94559 | 0.3 | 18.82658 | -2.2 | 0.67194 | -4.4 | 0.5882 | -7 |
| 10 | (1.00,0.05) | No | 18.89487 | -9.5 | 16.41411 | -14.7 | 0.57016 | -18.8 | 0.46423 | -26.6 |
| | | Yes | 20.80663 | -0.4 | 18.79957 | -2.3 | 0.67503 | -3.9 | 0.5996 | -5.2 |
| 11 | (1.00,0.10) | No | 17.568 | -15.9 | 14.8985 | -22.6 | 0.43027 | -38.8 | 0.27203 | -57 |
| | | Yes | 20.11792 | -3.7 | 17.97626 | -6.6 | 0.62461 | -11.1 | 0.52604 | -16.9 |
| 12 | (1.00,0.40) | No | 21.0745 | 0.9 | 19.39215 | 0.8 | 0.71071 | 1.1 | 0.65727 | 3.7 |
| | | Yes | 21.85928 | 4.4 | 20.19461 | 4.7 | 0.7337 | 4.2 | 0.68376 | 7.5 |
| 13 | (1.00,0.70) | No | 20.48647 | -1.9 | 18.76415 | -2.5 | 0.67241 | -4.3 | 0.60916 | -3.7 |
| | | Yes | 21.59631 | 3.3 | 19.94213 | 3.5 | 0.71816 | 2.2 | 0.66448 | 4.8 |
| 14 | (1.00,1.30) | No | 20.24277 | -3.1 | 18.34814 | -4.6 | 0.67416 | -4 | 0.60812 | -3.9 |
| | | Yes | 21.50951 | 2.9 | 19.7663 | 2.7 | 0.71966 | 2.4 | 0.66474 | 4.8 |
| 15 | (1.00,1.60) | No | 21.11148 | 1.1 | 19.33425 | 0.5 | 0.71201 | 1.3 | 0.65657 | 3.6 |
| | | Yes | 21.89409 | 4.8 | 20.18368 | 4.7 | 0.73519 | 4.4 | **0.68461** | 7.6 |
| 16 | (1.00,1.90) | No | 17.99903 | -13.8 | 15.03822 | -21.8 | 0.46759 | -33.4 | 0.29978 | -52.6 |
| | | Yes | 20.34618 | -2.6 | 18.04406 | -6.2 | 0.63809 | -9.2 | 0.53595 | -15.3 |
| 17 | (1.00,1.95) | No | 18.4333 | -11.8 | 15.75836 | -18.1 | 0.51124 | -27.2 | 0.37347 | -41 |
| | | Yes | 20.586 | -1.5 | 18.44235 | -4.2 | 0.65445 | -6.8 | 0.56461 | -10.8 |

Table 6 The results with FrScatNets on CelebA dataset. The best results are shown in bold. In the third column, "No" means un-fused image and "Yes" means fused image. Increased1, Increased2, Increased3, Increased4 mean the percentages of relative improvements on Train PSNR, Test PSNR, Train SSIM, Test SSIM of FrScatNets of various fractional orders ($α_1$, $α_2$) over conventional ScatNets, respectively.

| Index | ($α_1$, $α_2$) | Fusion or not? | Train PSNR | Increased 1 (%) | Test PSNR | Increased 2 (%) | Train SSIM | Increased3 (%) | Test SSIM | Increased4 (%) |
|---|---|---|---|---|---|---|---|---|---|---|
| 1 | (1.00,1.00) | No | 24.02802 | 0 | 21.99852 | 0 | 0.73765 | 0 | 0.70366 | 0 |
| 2 | (0.05,1.00) | No | 22.6989 | -5.5 | 20.46278 | -7 | 0.6951 | -5.8 | 0.65511 | -6.9 |
| | | Yes | 24.45131 | 1.7 | 22.49153 | 2.2 | 0.75149 | 1.8 | 0.72342 | 2.7 |
| 3 | (0.10,1.00) | No | 21.4913 | -10.6 | 19.09143 | -13.2 | 0.64422 | -12.7 | 0.58855 | -16.4 |
| | | Yes | 23.88528 | -0.6 | 21.84163 | -0.7 | 0.73355 | -0.6 | 0.69906 | -0.7 |
| 4 | (0.40,1.00) | No | 24.21105 | 0.8 | 22.25298 | 1.1 | 0.75231 | 1.9 | 0.72766 | 3.3 |
| | | Yes | 25.12589 | 4.4 | 23.24009 | 5.3 | 0.77408 | 4.7 | 0.75259 | 6.5 |
| 5 | (0.70,1.00) | No | 24.0699 | 0.2 | 22.23468 | 1.1 | 0.74503 | 1 | 0.7205 | 2.3 |
| | | Yes | 25.06707 | 4.1 | 23.24323 | 5.4 | 0.77171 | 4.4 | 0.75023 | 6.2 |
| 6 | (1.30,1.00) | No | 24.09127 | 0.3 | 22.22689 | 1 | 0.74631 | 1.2 | 0.72119 | 2.4 |
| | | Yes | 25.07891 | 4.2 | 23.24216 | 5.4 | 0.77179 | 4.4 | 0.75012 | 6.2 |
| 7 | (1.60,1.00) | No | **24.22473** | 0.8 | **22.31411** | 1.4 | **0.75554** | 2.4 | **0.73113** | 3.8 |
| | | Yes | **25.13193** | 4.4 | **23.27425** | 5.5 | **0.77591** | 4.9 | **0.75456** | 6.7 |
| 8 | (1.90,1.00) | No | 21.80923 | -9.2 | 19.36504 | -12 | 0.65626 | -11 | 0.60019 | -14.7 |
| | | Yes | 24.04234 | 0.1 | 21.97893 | -0.1 | 0.73795 | 0.04 | 0.70348 | -0.02 |
| 9 | (1.95,1.00) | No | 22.66739 | -5.7 | 20.28151 | -7.8 | 0.69154 | -6.3 | 0.64782 | -7.9 |
| | | Yes | 24.45157 | 1.7 | 22.41367 | 1.9 | 0.75022 | 1.7 | 0.7206 | 2.4 |
| 10 | (1.00,0.05) | No | 22.62422 | -5.8 | 20.57329 | -6.5 | 0.68594 | -7 | 0.64726 | -8 |
| | | Yes | 24.39819 | 1.5 | 22.5255 | 2.3 | 0.74739 | 1.3 | 0.72007 | 2.3 |
| 11 | (1.00,0.10) | No | 21.29682 | -11.4 | 18.96768 | -13.8 | 0.63095 | -14.5 | 0.57726 | -18 |
| | | Yes | 23.7895 | -1 | 21.77061 | -1 | 0.72801 | -1.3 | 0.69406 | -1.4 |
| 12 | (1.00,0.40) | No | 24.253 | 0.9 | 22.3492 | 1.6 | 0.74327 | 0.8 | 0.71779 | 2 |
| | | Yes | 25.0994 | 4.3 | 23.2497 | 5.4 | 0.76872 | 4 | 0.74674 | 5.8 |
| 13 | (1.00,0.70) | No | 23.98679 | -0.2 | 22.08526 | 0.4 | 0.73254 | -0.7 | 0.70602 | 0.3 |
| | | Yes | 25.02702 | 4 | 23.16735 | 5 | 0.76486 | 3.6 | 0.74247 | 5.2 |
| 14 | (1.00,1.30) | No | 23.97447 | -0.2 | 22.06954 | 0.3 | 0.7337 | -0.5 | 0.70742 | 0.5 |
| | | Yes | 25.03175 | 4 | 23.16357 | 5 | 0.76537 | 3.6 | 0.73201 | 3.9 |
| 15 | (1.00,1.60) | No | 24.20664 | 0.7 | 22.33477 | 1.5 | 0.74501 | 1 | 0.71988 | 2.3 |
| | | Yes | 25.08629 | 4.2 | 23.24543 | 5.4 | 0.76947 | 4.1 | 0.74762 | 5.9 |
| 16 | (1.00,1.90) | No | 21.9754 | -8.5 | 19.45485 | -11.6 | 0.65881 | -10.7 | 0.6037 | -14.2 |
| | | Yes | 24.10325 | 0.3 | 22.00569 | 0.03 | 0.73737 | -0.03 | 0.70314 | -0.1 |
| 17 | (1.00,1.95) | No | 22.55721 | -6.1 | 20.32281 | -7.6 | 0.6814 | -7.6 | 0.63903 | -9.2 |
| | | Yes | 24.41815 | 1.6 | 22.44185 | 2 | 0.74596 | 1.1 | 0.71717 | 1.9 |

The generative images on CIFAR-10 dataset using FrScatNets with $(\alpha_1, \alpha_2)$=(0.40, 1.00) and $(\alpha_1, \alpha_2)$=(1.00, 1.00) are shown in Fig. 12. The generative images on CelebA dataset using FrScatNets with $(\alpha_1, \alpha_2)$=(1.60, 1.00) and $(\alpha_1, \alpha_2)$=(1.00, 1.00) are shown in Fig. 13.

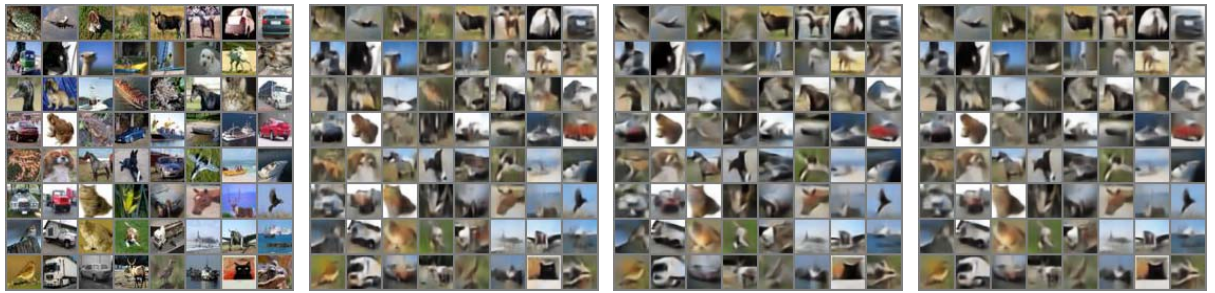

(a)  (b)  (c)  (d)

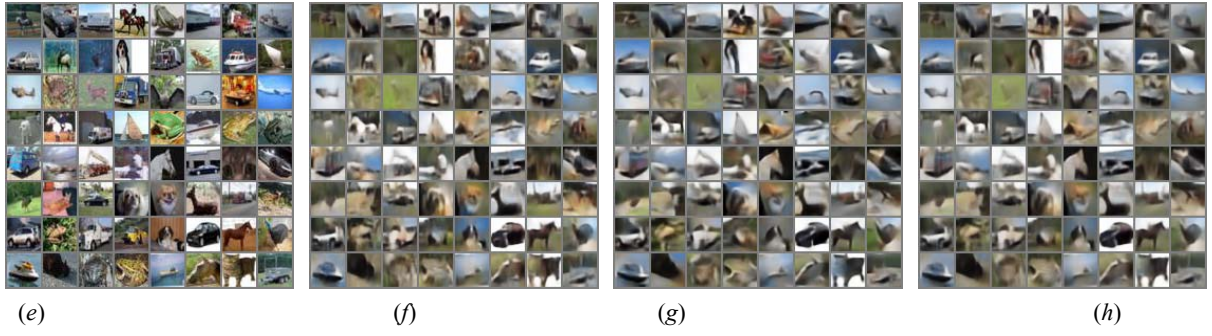

(e)          (f)          (g)          (h)

Fig. 12 The generative images on CIFAR-10 dataset using FrScatNets embedding. (*a*) Original training images; (*b*) Generative training images using FrScatNets with $(\alpha_1, \alpha_2)=(0.40, 1.00)$; (*c*) Generative training images using FrScatNets with $(\alpha_1, \alpha_2)=(1.00, 1.00)$; (*d*) Fused training image using FrScatNets with $(\alpha_1, \alpha_2)=(0.40, 1.00)$ and $(\alpha_1, \alpha_2)=(1.00, 1.00)$; (*e*) Original testing images; (*f*) Generative testing images using FrScatNets with $(\alpha_1, \alpha_2)=(0.40, 1.00)$; (*g*) Generative testing images using FrScatNets with $(\alpha_1, \alpha_2)=(1.00, 1.00)$; (*h*) Fused tesing image using FrScatNets with $(\alpha_1, \alpha_2)=(0.40, 1.00)$ and $(\alpha_1, \alpha_2)=(1.00, 1.00)$.

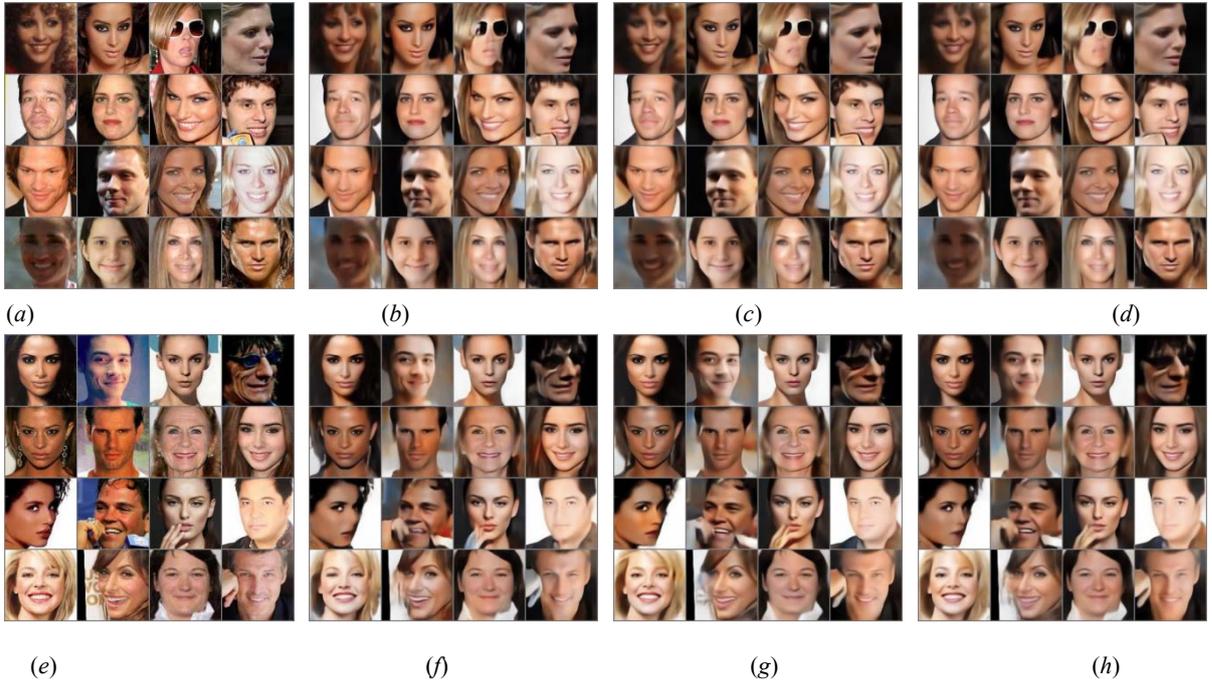

(a)          (b)          (c)          (d)

(e)          (f)          (g)          (h)

Fig. 13 The generative images on CelebA dataset using FrScatNets embedding. (*a*) Original training images; (*b*) Generative training images using FrScatNets with $(\alpha_1, \alpha_2)=(1.60, 1.00)$; (*c*) Generative training images using FrScatNets with $(\alpha_1, \alpha_2)=(1.00, 1.00)$; (*d*) Fused training image using FrScatNets with $(\alpha_1, \alpha_2)=(1.60, 1.00)$ and $(\alpha_1, \alpha_2)=(1.00, 1.00)$; (*e*) Original testing images; (*f*) Generative testing images using FrScatNets with $(\alpha_1, \alpha_2)=(1.60, 1.00)$; (*g*) Generative testing images using FrScatNets with $(\alpha_1, \alpha_2)=(1.00, 1.00)$; (*h*) Fused tesing image using FrScatNets with $(\alpha_1, \alpha_2)=(1.60, 1.00)$ and $(\alpha_1, \alpha_2)=(1.00, 1.00)$.

## 4.3 Image generative results with image fusion

In this subsection, we explore the impact of image fusion on the quality of generated images using the framework of GFRSNs shown in Fig. 5. The other parameter settings of FrScatNets are shown in

Table 2 The core parameters of FrScatNet with and without feature dimensional reduction.

Table 2. We choose the $L_1$ loss function in (15) and train the generator by Adam optimizer using the default hyperparameters.

The generative images from FrScatNets with different fractional orders ($\alpha_1$, $\alpha_2$), where $\alpha_1$ and $\alpha_2$ are not 1.00 at the same time, are fused with the generative image from conventional ScatNets, that is, FrScatNets with fractional orders ($\alpha_1$, $\alpha_2$)=(1.00, 1.00). All fused images are achieved by the average method shown in Eq. (17) and we choose $\lambda$=0.5. The PSNR and SSIM results of fused image on CIFAR-10 dataset are shown in Table 5 and on CelebA dataset are shown in Table 6. Note that the results are shown in the row where the "Fusion or not?" column is "Yes" in Table 5 and Table 6. As we can see from the two tables, the results of PSNR and SSIM of the fused image are generally better than the unfused image from FrScatNets with different fractional orders ($\alpha_1$, $\alpha_2$), where $\alpha_1$ and $\alpha_2$ are not 1.00 at the same time. For example, when the generative image from FrScatNets with ($\alpha_1$, $\alpha_2$)= (0.40, 1.00) are fused with generative image from ScatNets, the Test PSNR and the Test SSIM are increased from 19.52272 and 0.65852 to 20.23685 and 0.68317, respectively, on CIFAR-10 dataset. The results are also better than the ScatNets based GFRSRNs, whose Test PSNR and the Test SSIM are 19.24235 and 0.63264, respectively. When the generative image from FrScatNets with ($\alpha_1$, $\alpha_2$)= (1.60, 1.00) are fused with generative image from ScatNets, the Test PSNR and the Test SSIM are increased from 22.31411 and 0.73113 to 23.27425 and 0.75456, respectively, on CelebA dataset. The results are also better than the ScatNets based GFRSRNs, whose Test PSNR and the Test SSIM are 21.99852 and 0.70366, respectively. The fused images on CIFAR-10 dataset is shown in Fig. 13 (d), (h) and on CelebA dataset is shown in Fig. 14 (d), (h), respectively.

## 5. Conclusion

This paper proposes Generative Fractional Scattering Networks (GFRSNs), which use fractional wavelet scattering networks (FrScatNets) as the encoder to obtain the features (or FrScatNet embeddings) and use deconvolutional neural networks as the decoder to generate the image. Additionally, this paper develops a new Feature-Map Fusion (FMF) method to reduce the dimension of FrScatNet embeddings. The impact of image fusion is also discussed in the paper. The experimental results on CIFAR-10 dataset and on CelebA dataset show that the proposed GFRSNs can obtain better generated images than the original GSNs in testing dataset.

**ACKNOWLEDGEMENT**

This work was funded by the National Natural Science Foundation of China under Grants 61876037, 31800825, 61871117, 61871124, 61773117, 61872079, and in part by the National Science and Technology Major Project of the Ministry of Science and Technology of China under Grant 2018ZX10201002-003, and in part by the Short-Term Recruitment Program of Foreign Experts under Grant WQ20163200398, and in part by INSERM under the Grant call IAL. We thank the Big Data Computing Center of Southeast University for providing the facility support on the numerical calculations in this paper.